\documentclass[10pt,twocolumn,letterpaper]{article}

\usepackage{cvpr}
\usepackage{times}
\usepackage{epsfig}
\usepackage{graphicx}
\usepackage{amsmath}
\usepackage{amssymb}
\usepackage{stackengine}
\usepackage{multirow}

\cvprfinalcopy 

\def\cvprPaperID{1607} 

\ifcvprfinal\fi
\begin{document}

\title{Scalable High Quality Object Detection}

\author{
Christian Szegedy\\
Google Inc.\\
1600 Amphitheatre Pkwy, Mountain View, CA\\
{\tt\small szegedy@google.com}
\and
Scott Reed\\
University of Michigan \\
{\tt\small reedscot@umich.edu}
\and
Dumitru Erhan\\
{\tt\small dumitru@google.com}
\and
Dragomir Anguelov\\
{\tt\small dragomir@google.com}
\and
Sergey Ioffe\\
{\tt\small sioffe@google.com}
}

\maketitle

\begin{abstract}
Current high-quality object detection approaches use the same scheme:
salience-based object proposal methods followed by post-classification using
deep convolutional features.  This spurred recent research in improving object
proposal methods
~\cite{PABMM2015,zitnick2014edge,manen2013prime,hosang2014good,cheng2014bing}.
However, domain agnostic proposal generation has the principal drawback that the
proposals come unranked or with very weak ranking, making it hard to trade-off
quality for running time.
Also, it raises the more fundamental question of whether high-quality proposal
generation requires careful engineering or can be derived just from data alone.
We demonstrate that learning-based proposal methods can effectively match the
performance of hand-engineered methods while allowing for very efficient runtime-quality trade-offs.
Using our new multi-scale convolutional MultiBox (MSC-MultiBox)
approach, we substantially advance the state-of-the-art on
the ILSVRC 2014 detection challenge data
set, with $0.5$ mAP for a single model and $0.52$ mAP for
an ensemble of two models. MSC-Multibox significantly improves the proposal
quality over its predecessor Multibox~\cite{erhan2014scalable}
method: AP increases from $0.42$ to $0.53$ for the ILSVRC detection challenge.
Finally, we demonstrate improved bounding-box recall compared to
Multiscale Combinatorial Grouping~\cite{PABMM2015} with less proposals on the
Microsoft-COCO~\cite{lin2014microsoft} data set.
\end{abstract}
\vspace{-0.25in}

\section{Introduction}
After the dramatic improvements in object detection demonstrated by Girshick et al.~\cite{girshick2014rcnn}, most of the current state of the art approaches, including the top performing entries~\cite{szegedy2014going,ouyang2014deepid,simonyan2014very} of the 2014 Imagenet object detection competition~\cite{russakovsky2014imagenet}, make use of salience-based object localization, in particular Selective Search~\cite{uijlings2013selective} followed by some post-classification method using features from a deep convolutional network.

Given the fact that the best salience-based methods can reach up to 95\% coverage of all objects at 0.5 overlap threshold on the detection challenge validation set, it is tempting to focus on improving the post-classification ranking alone while considering the proposal generation part to be solved.
However, this might be a premature conclusion: a better way of ranking the proposals is to cut down their number at generation time already. In the ideal case,
we will be able to achieve high coverage with very few proposals. This can improve not only the running time but also the quality, because the post-classification
stage would need to handle fewer potential false positives. Furthermore, a strong proposal ranking function provides a way to balance recall versus running-time in a
simple, consistent manner by just selecting appropriate thresholds: use a high threshold for use cases where speed is essential,
and a low threshold when quality matters most.

Motivated by the fact that hand-engineered features are getting replaced by higher-quality deep neural network features for image classification~\cite{lecun1995learning,krizhevsky2012imagenet,szegedy2014going}, we show that the same trend holds for proposal generation. In Section~\ref{sec:proposalquality} we demonstrate that our purely learned
proposal method closely rivals salience-based methods in performance, at a significantly lower computational cost. Furthermore, the ability to directly
learn region proposal methods is a key advantage as it is easy to adapt the model to new domains such as medical or aerial imaging or to specific use cases, such as
recognizing only certain objects. In contrast, hand-engineered proposal methods are typically tuned for natural objects with clear segmentation, but do less
well in domains where the distinction between objects needs more subtle cues and cannot return proposals only for objects of interest.

Our work builds upon the MultiBox approach presented in~\cite{erhan2014scalable}, which was an earlier attempt to learn a proposal generation model but was
never directly competitive with the best expert-engineered alternatives. We demonstrate that switching to the latest Inception~\cite{szegedy2015rethinking}-style
architecture and utilizing multi-scale convolutional predictors of bounding box shape and confidence, in combination with an Inception-based
post-classification model significantly improves the proposal quality and the final object detection quality.
Combining this with a simple but efficient contextual model, we end up with a single system that scales to a variety of use cases
from real-time to very high-quality detection and achieves a new state of the art result on the ImageNet detection challenge.

In summary, the main contributions of our approach are:
\begin{itemize}
\item Improved network architecture for bounding box generation, including multi-scale convolutional bounding box predictors.
\item Integration of a context model during post-classification, which improves performance.
\item 200 classes detection at $0.45$ mAP with $15$ proposals per image generated by our box proposal method.
\item $0.50$ mAP with a single model and $0.52$ with an ensemble of three post-classifiers and two MultiBox
proposal generators.
\end{itemize}
Additionally, in Sec.~\ref{sec:results} we analyze the effect of the various components of the MSC-Multibox model.

\section{Related Work}

The previous state-of-the-art paradigm in detection is to use part-based models~\cite{fischler1973representation,felzenszwalb2010object} such as Deformable Part Models (DPMs). Sadeghi and Forsyth~\cite{sadeghi201430hz} developed a framework with several configurable runtime-quality trade-offs and demonstrate real-time detection using DPMs on the PASCAL 2007 detection data.

Deep neural network architectures with repeated convolution and pooling layers~\cite{fukushima1979neural,lecun1995learning} have more recently become the dominant approach for large-scale and high-quality recognition and detection.
Szegedy et al.~\cite{szegedy2013deep} used deep neural networks for object detection formulated as a regression onto bounding box masks. Sermanet et al.~\cite{sermanet2013overfeat} developed a multi-scale sliding window approach using deep neural networks, winning the ILSVRC2013 localization competition.

The original work on MultiBox~\cite{erhan2014scalable} also used deep networks, but focused on increasing efficiency and scalability. Instead of producing bounding box masks, the MultiBox approach directly produces bounding box coordinates, and avoids linear scaling in the number of classes by making class-agnostic region proposals.
%
In our current work (detailing improvements to MultiBox) we demonstrate greatly increased recall of object locations by increasing the number of \emph{potential} proposals with a fixed budget of evaluated proposals.
We also demonstrate improvements to the training strategy and underlying network architecture that yield state-of-the-art performance.

Other recent works have also attempted to improve the scalability of the now-predominant R-CNN detection framework~\cite{girshick2014rcnn}.
He et al. proposed Spatial Pyramid Pooling~\cite{he2014spatial} (SPP), which engineers robustness to aspect-ratio variation into the network.
They also improve the speed of evaluating Selective Search proposals by classifying mid-level CNN features (generated from a single feed-forward pass) rather than pushing all image crops through a full CNN.
They report roughly two orders of magnitude ($\sim 100$x) speedup over R-CNN using their method.

Compared to the SPP approach, we show a comparable efficiency improvement by drastically reducing the number and improving the quality of region proposals via our MultiBox network, which also associates a confidence score to each proposal.
Architectural changes to the underlying network and contextual
post-classification were the main factors in reaching high quality.
%
%
We emphasize that MultiBox and SPP are complementary in the sense that spatial pyramid pooling can be added to the underlying ConvNet if desired, and post-classification of proposals can be sped up in the same way with no change to the MultiBox objective.

Another way in which efficiency of detection methods can be improved is by
unifying the detection and classification models, reusing as much computation
as possible and in the process abandoning the idea of data-independent region
proposals. An example of such an approach is Pinheiro et
al.~\cite{pinheiro2015learning}'s work, who propose a convolutional neural
network model with two branches: one that can generate class-agnostic
$segmentation$ masks, and second branch predicting the likelihood of a given
patch being centered on an object.  Inference is efficient since the model is
applied convolutionally on an image and one can get the class scores and
segmentation masks using a single model.

The YOLO approach by Redmon et al.~\cite{redmon2015you} is similar to it, in
that it uses a single network to predict bounding boxes and class
probabilities, in an end to end network. The difference is that it
divides the input image into a grid of cells and predicts the coordinates and
confidences of objects contained in the cells.  This approach is fast, but
limited in that each grid cell can only contain one object by construction,
with the grid being quite coarse. It is also unclear to which extent these
results can translate to good performance on data sets with significantly more
objects, such as the ILSVRC detection challenge.

Faster R-CNN~\cite{ren15fasterrcnn} is a technique that merges the
convolutional features of the full-image network with the detection network,
thereby simultaneously predicting boxes and objectness scores. The detection
network--called the Region Proposal Network (RPN)--is trained end to end in an
alternating fashion with the Fast R-CNN network; its objective is to produce
good region proposals. The RPN is thus quite similar to the
Multibox approach described in this paper: the two approaches have been
co-developed at the same time. The biggest similarity is the usage of priors
(called ``anchors’’ in the Fast R-CNN work~\cite{girshickICCV15fastrcnn}) that are designed to be translation
invariant and that are predicted from the top layer feature map.  Our
multiscale priors are different in that we use multiple tapering layers, while
the Fast R-CNN approach is predicting boxes of many scales from a single
feature map. The other differences include the fact that in our approach the
confidences are class-agnostic, and we used different box regression and
classification losses.  Notably, we also use radically different network
architectures, with parts designed specifically to overcome the shortcomings of
networks designed for classification. Finally, we argue that our two-stage setup
scales to a higher number of classes well: the Faster R-CNN work uses many thousands
of priors and scaling that approach to thousands of classes is not obvious.
It would ultimately be interesting to
disentangle which of these differences are important, by comparing the two
methods on the same evaluation set.

\section{Model}
\subsection{Background: MultiBox objective}
%
%

In order to describe the changes to~\cite{erhan2014scalable}, let us revisit the basic tenets of the MultiBox method.
The fundamental idea is to train a convolutional network that outputs the coordinates of the object bounding boxes directly.
However, this is just half of the story, since we would also like to rank the proposals by their likelihood of being
an accurate bounding box for an object of interest.
In order to achieve this, the MultiBox loss is the weighted sum of the following two losses:
\begin{itemize}
\item \textbf{Confidence}: a logistic loss on the estimates of a proposal corresponding to an object of interest.
\item \textbf{Location}: a loss corresponding to some similarity measure between the objects and the closest matching object box predictions. By default we used L2 distance.
\end{itemize}
The network is an improved Inception-style~\cite{szegedy2015rethinking} convolutional network, followed by a structured output module producing a set of bounding box coordinates and confidence scores. In the original MultiBox solution, the predictors were fully connected to the top layer of the network. Here we propose a multi-scale convolutional architecture described below.

Let $l_{i} \in \mathbb{R}^{4}$ be the $i$-th set of predicted box coordinates for an image, and let $g_{j} \in \mathbb{R}^{4}$ be the $j$-th ground-truth box coordinates.
At training time, for each image, we perform a bipartite matching between predictions and ground-truth boxes.
We denote $x_{ij} = 1$ to indicate that the $i$-th prediction is matched to the $j$-th ground-truth, and $x_{ij} = 0$ otherwise. Note that $x$ is constrained so that $\sum_{i}x_{ij} = 1$.
Given a matching between predictions and groundtruth, the location loss term can be written as
\begin{align}
	F_{loc}(x,l,g) = \dfrac{1}{2}\sum_{i,j}x_{ij}||l_{i} - g_{j}||^{2}_{2}.
\end{align}
Given the predicted scores $c_i$, the confidence loss term can
be written as follows:
\begin{align}
	F_{conf}(x,c) =  - & \sum_{i,j} x_{ij}\log(c_{i}) -\\
			   & \sum_{i}(1-\sum_{j}x_{ij})\log(1-c_{i})\nonumber
\end{align}
The overall objective is a weighted sum of both terms
\begin{align}
	F(x,c,l,g) = F_{conf}(x,c) + \alpha F_{loc}(x,l,g)
\end{align}
We train the network with stochastic gradient descent. For each training example  with ground truth $g$ and network output $(c,l)$ we compute the matching $x^*$ by picking the minimizer of the loss:
\begin{eqnarray}
	x^{*} & = & \underset{x}{\text{arg min }}F(x,c,l,g) \\
	\text{such that } &  & x_{ij} \in \{ 0,1 \} \text{ , } \sum_{i}x_{ij} = 1 \nonumber
\end{eqnarray}
and update the network parameters following the gradient evaluted at the matching $x^{*}$ that was found.

\subsection{Convolutional Priors}
\begin{figure}
\centering
\includegraphics[width=\linewidth]{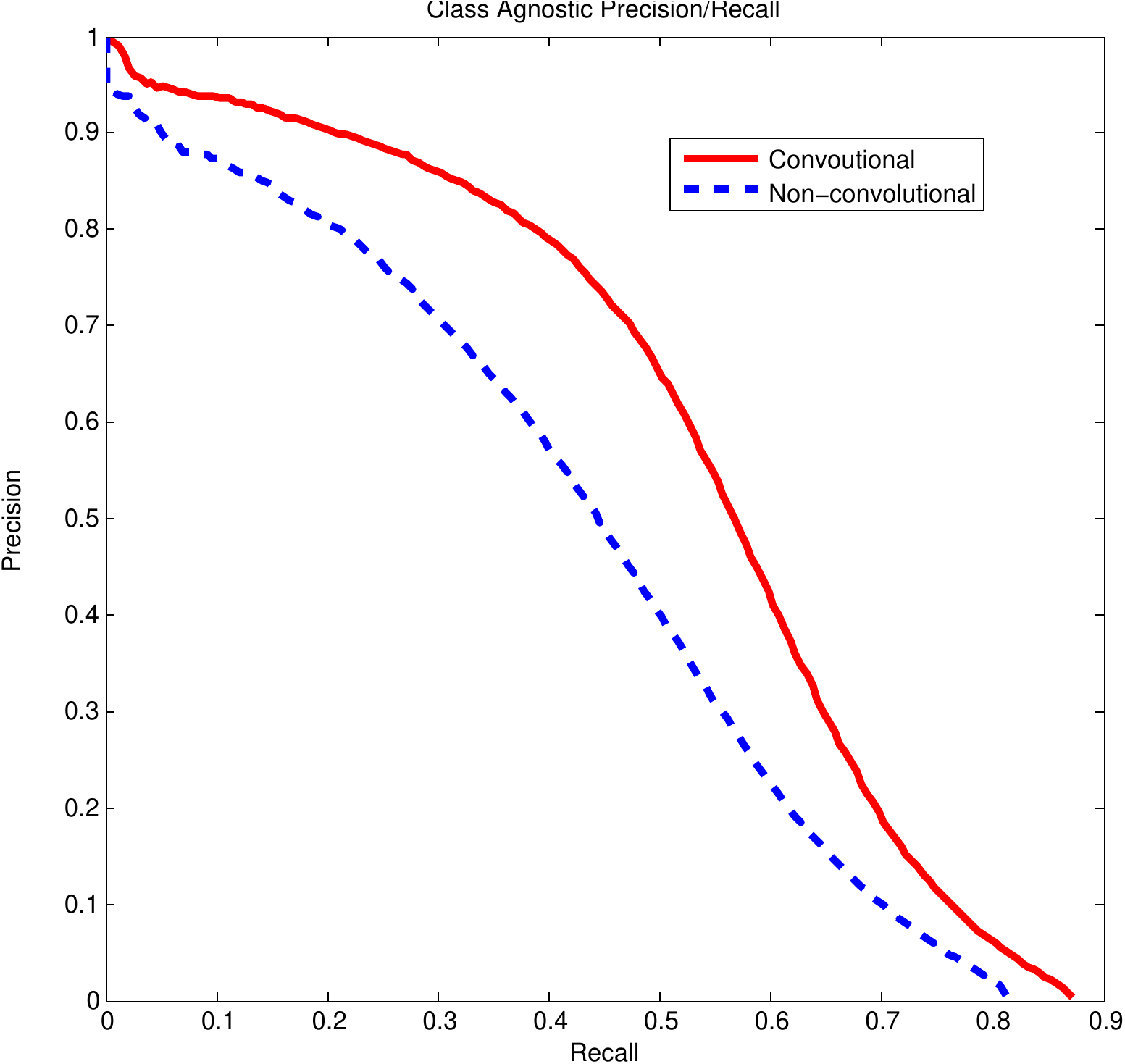}
\caption{Class-agnostic Precision-Recall at IOU threshold $0.5$ of the MultiBox trained with convolutional vs. non-convolutional priors. The (class-agnostic) average precision goes up from $0.417$ to $0.529$. }
\label{fig:priors}
\end{figure}

The MultiBox~\cite{erhan2014scalable} setup is to predict locations
(the five coordinates) and confidences for a constant number of boxes.
We call the associated outputs of the network ``slots'': each
slot corresponds to one predicted proposal. However, these proposals
might be low confidence, in which case the network predicts that the
associated box does not correspond to any object on the image.
Our goal is to maximize the coverage of the high-confidence predictions.
Our network is an ``objectness'' detector, but our notion
of what constitutes an object depends on the task we try to tackle.

A crucial detail of our approach is that we do not let the proposals
free-float, but impose diversity by introducing a prior for each box
output slot of the network. Let us assume the our network predicts $k$ boxes,
together with their confidences, then each of those output slots will be
associated with a prior rectangle $p_i$.
These rectangles are computed before training the network
in a way that matches the distribution of object boxes in the training set.
Our goal is to maximize the expected coverage of this constant set of
priors at a given Jaccard (IOU) overlap threshold $t=0.5$.
In \cite{erhan2014scalable}, the goal was to maximize the expected overlap between
each ground-truth object box and the best matching prior.
Here we try to find a set of priors to optimize
$E([min_{\{p_i\}}(IOU(p_i, b_i)) > t])$, where $b_i$ are matching groundtruth bounding boxes.
Intuitively, we can say that we want
the best proposal generation method that is independent of the image pixels
and has the maximum coverage at a given overlap threshold $t$ ($0.5$
in our case).

\begin{figure*}
\centering
\includegraphics[width=\linewidth]{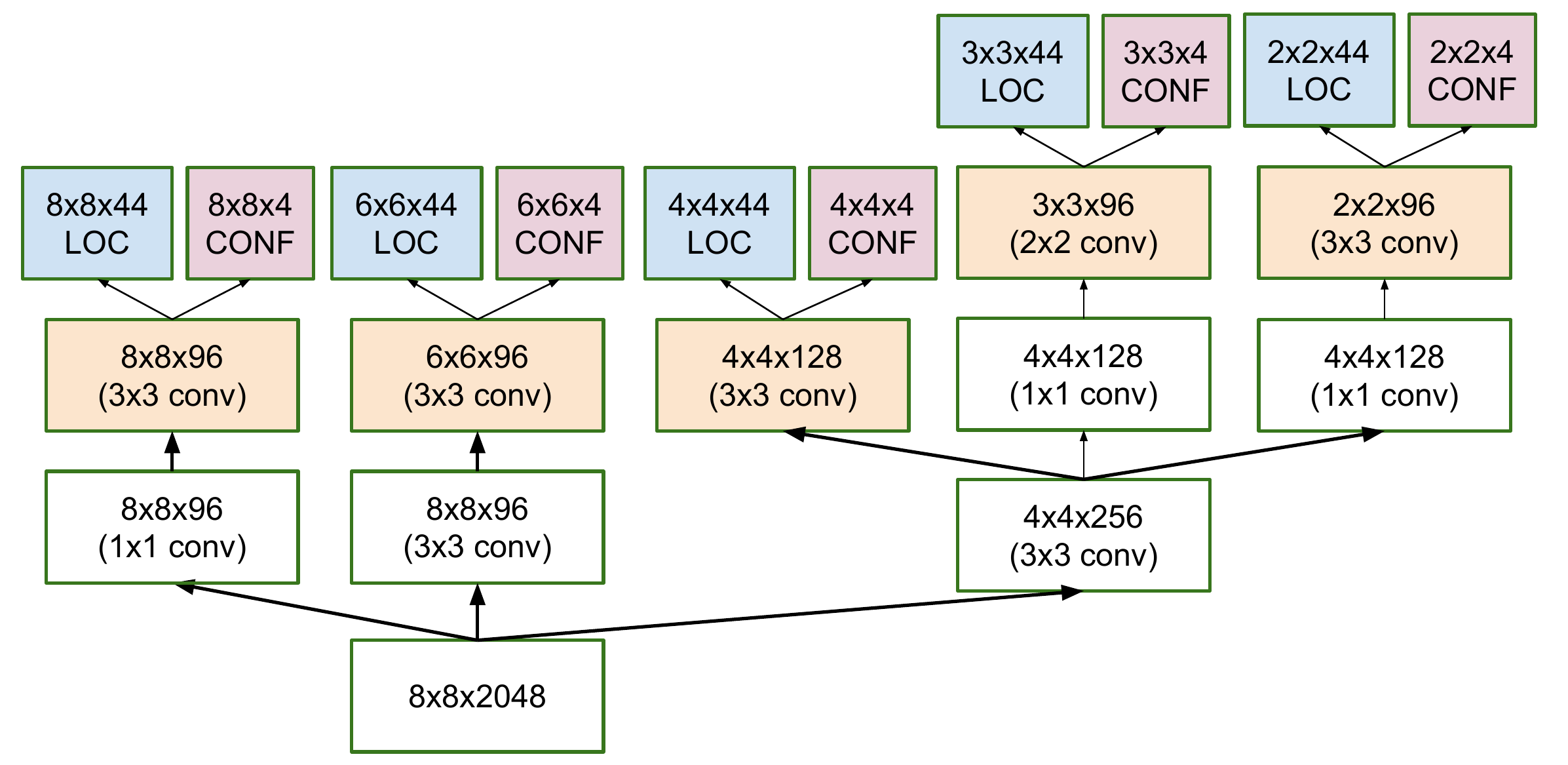}
\caption{An illustration of the multi-scale convolutional prediction of the locations and confidences for MultiBox.}
\label{fig:multiscale}
\end{figure*}

As in Multibox~\cite{erhan2014scalable}, the bounding boxes $l_i$ predicted
by slot $i$ of the network will be interpreted with respect to prior $p_i$.
That is, we are regressing toward $g_j - p_i$ where $g$ is the groundtruth
box minimizing $\|g_j - p_i\|$ and at inference time if the network
outputs $l_i'$ for slot $i$, the predicted box $l_i$ will be set to $l_i' + p_i$.
Erhan \emph{et~al.}~\cite{erhan2014scalable} took a similar approach, but they tried to
maximize the expected overlap as opposed to the coverage.
However, it is a highly non-convex objective function, so they needed
to resort to the heuristic of performing $k$-means clustering of
the ground-truth object boxes of the training set objects and took
the $k$-means centroids as priors.
Here, we are taking a different approach that is closely related to the approach
of Faster R-CNN~\cite{ren15fasterrcnn} and exploits the expected
translation invariance of the object locations in the data set.
The priors are assumed to lie on grids with grid lines parallel
to the image boundary. Formally, we assume that our set $P$ of prior
boxes is the union of boxes placed regularly on those grids.
\begin{align}
  P = \bigcup_q (G_q + t_q),
\end{align}
where $G_q=\delta_q\{1,\dots m_q\}\times\delta_q\{1\dots m_q\}$ is a regular
two dimensional grid and $t_q$ is the template box displaced by the grid
and $m_q$ denotes the grid resolution. In our setup we have set
$\delta_q=\frac{1}{m_q + 1}$.
In addition to the $8\times 8$ top layer of our base network,
we add a prediction tree to our network as depicted in Fig.~\ref{fig:multiscale}.
We have a dedicated layer for producing prediction locations and scores for
each of the $8\times 8$, $6\times 6$, a $4\times 4$, a $3\times 3$
a $2\times 2$ and a $1\times 1$ grids (the $1\times 1$ grid is created by applying
average pooling on the $8\times 8$ top base network layer).
Each tile of each grid but the $1\times 1$ is responsible for predicting $11$ outputs with
priors of different aspect ratios. The top $1\times 1$ grid is used for
predicting the single largest prior.
This way we end up using
\begin{align}
1 + 11 \times (8 \times 8 + 6 \times 6 + 4 \times 4 + 3 \times 3 + 2 \times 2)) =1420 \nonumber
\end{align}
priors. Each of these priors is associated with one location output
slot and its associated confidence output slot of the network. The outputs
are emitted by the $LOC$ and $CONF$ layers as shown in Fig.~\ref{fig:multiscale}.

\subsection{Training with missing positive labels}
In large-scale data sets such as ImageNet, there are many missing true-positive labels.
In the confidence term of the MultiBox training objective, a large loss will be incurred if the model assigns a high confidence to a true positive object in the image that is missing a label.
We hypothesize that the dissonance caused by missing or noisy training data may encourage the model to be overly conservative in its predictions and thereby reduce the recall of MultiBox proposals. To deal with the issue, we adopted the ``hard bootstrapping'' approach of Reed et al.~\cite{reed2014noisy}.

Training with this method is equivalent to reformulating the confidence objective as follows:
\begin{align}
	F_{bootstrap}&(x,c) =  -\sum_{i}1_{ \{ i \notin topL(c) \} } \\
			& ( \sum_{j}x_{ij}\log c_{i} + (1 - \sum_{j}x_{ij})\log(1 - c_{i}) ), \nonumber
\end{align}
where $topL(c)$ is the set of indices into the top-$L$ most confident predictions. In practice, we precompute $topL(c)$ for every image within a batch before computing the gradients. The learning iterates between ``generating data'' according to the previous model state, and then updating the model based on the augmented data. In our experiments we initialized the network with networks pre-trained with no bootstrapping, and then fine-tuned on $F_{bootstrap}$.
%

\subsection{MultiBox network architecture}
For both the MultiBox localizer model and the post-classifier, we have been using new variants of the Inception architecture as described in~\cite{szegedy2015rethinking}. This is a 42 layers deep convolutional network over a $299\times 299$ receptive field, containing over 130 layers. We are using the top $8\times 8\times 2048$ convolutional layer as described earlier. The extra side heads are removed for simplicity. The exact architecture topology is given in the supplementary {\tt model.txt} file that can be downloaded together with the source file of this paper. Also we employ the spatially sensitive grid-size reduction technique as depicted in Figure~\ref{fig:hybridreduction}.

\subsection{Post-classification}
MSC-MultiBox can be used in two ways: as a one-shot detector that produces object locations and confidences, or as a class-agnostic localizer providing region proposals to a post-classifier.
However, in the high-quality regime, it is essential to zoom into the actual object proposals and perform an extra post-classification step to maximize performance.
%
%
When used in this setting, an additional post-classification step is necessary.
Again, for this use case we utilize the Inception architecture from~\cite{szegedy2015rethinking}.
%

\subsection{Post-classifier architecture improvements}
As a motivation for designing a new network architecture, we noted that the
post-classifier network not only needs to produce the
correct label for each class, but it also needs to decide whether the object
overlaps the crop occupying the center part of the receptive field. (We
follow the cropping methodology of the R-CNN~\cite{girshick2014rcnn} paper.)
This requires the network to be spatially sensitive.

\begin{figure}
\centering
\includegraphics[width=\linewidth]{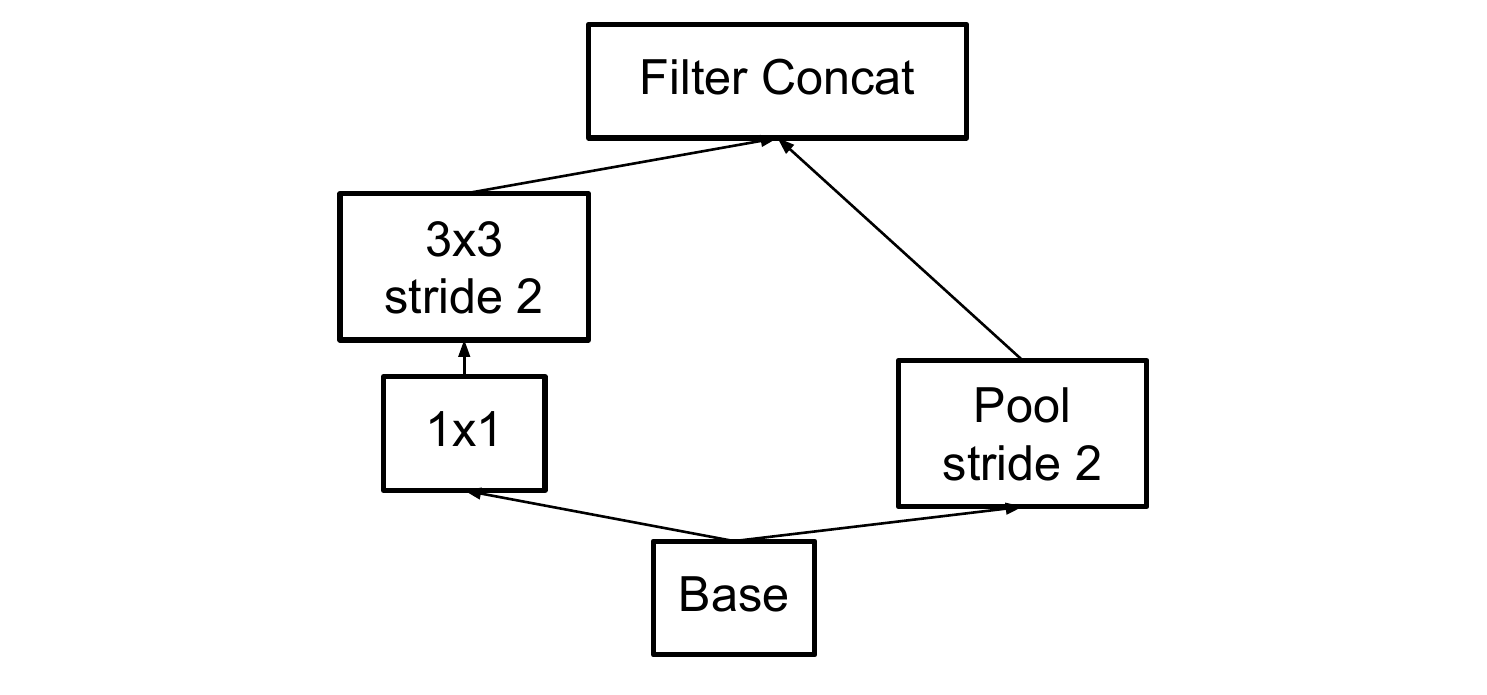}
\caption{Inception module that avoids using a pooling layer alone
  to do the grid reduction. The stride $2$ convolution can preserve
  the geometric information with less overhead.}
\label{fig:hybridreduction}
\end{figure}

We hypothesized that the large pooling layers of traditional network
architectures -- which are also inherited by the
Inception~\cite{szegedy2014going} architecture -- might be detrimental for
accurately predicting spatial information. This leads to the construction of
a variant of the Inception network, where in parallel to the
large pooling layers~\cite{weng1992cresceptron} stride-$2$ convolutions
are used in the Inception modules when reducing the grid size. This is depicted
in Fig.~\ref{fig:hybridreduction}.

\subsection{Context Modeling}
\begin{figure}
\centering
\includegraphics[width=0.6\linewidth]{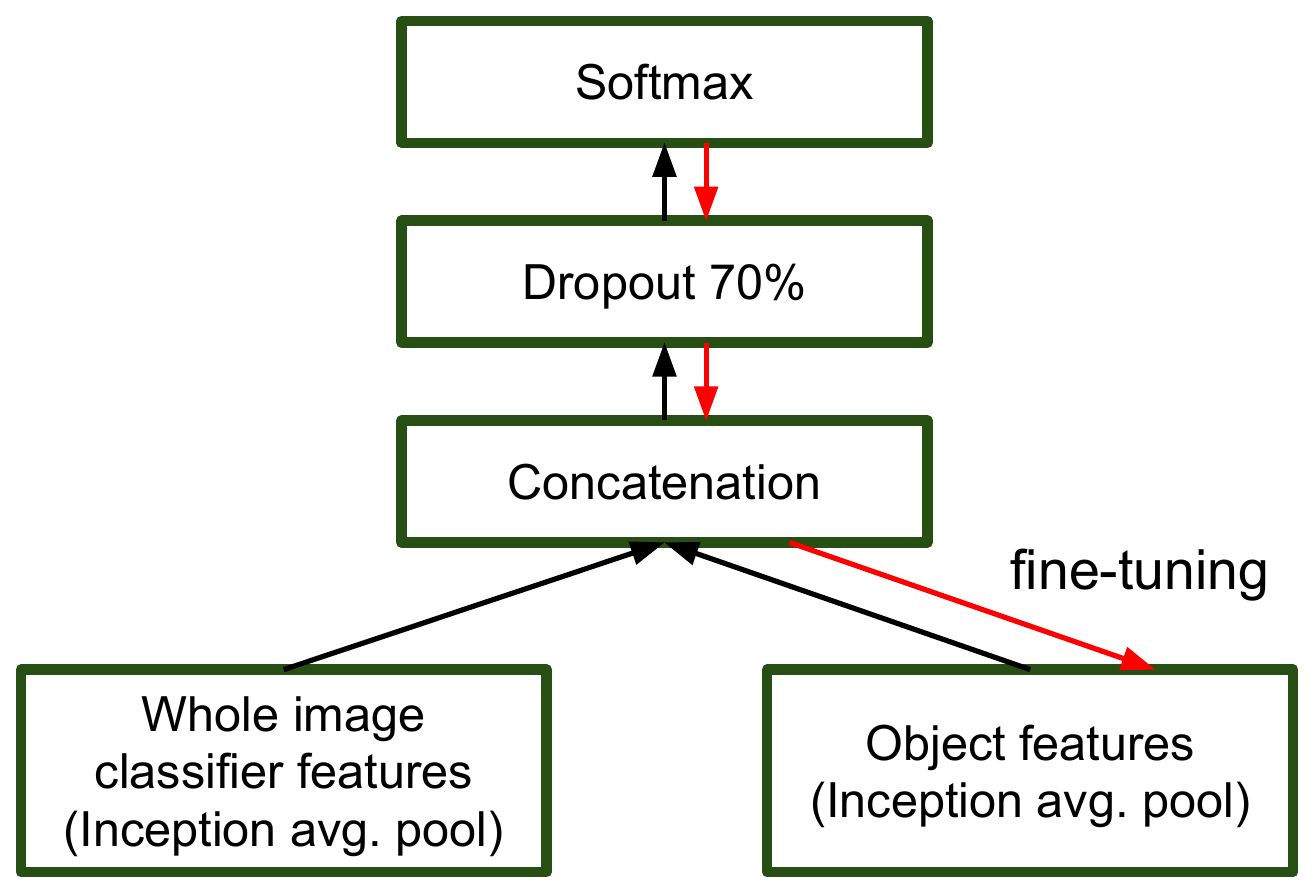}
\caption{Scheme of the combiner architecture. Note, that in our setting fine-tuning was performed only for the object-feature network, shown in red above.}
\label{fig:combiner}
\end{figure}
It is known that the global context can be useful when making predictions for local image regions.
Most high-performing detectors use elaborate schemes to update scores or take whole-image classification into account.
%
Instead of working with scores, we just concatenate the whole image features
with the object features, where the feature vector is taken from the topmost
layer before the classifier. See Fig.~\ref{fig:combiner}.

%
Note, however that two separate models are used for the context and object features and they don't share weights.

The context classification network is trained first with the logistic objective, meaning that we have a separate logistic classifier for each class and the sum of their losses is used as the total objective of the whole network.
We do not use the classifier output of the context network at object
proposal evaluation time. The combiner network in fig~\ref{fig:combiner} is
trained in a second step after whole image features have been extracted.
The combiner is uses a softmax classifier, since each bounding box can only have
a single class. A designated ``background'' class is used for crops that
don't overlap any of the objects with at least $0.5$ intersection over union
(IOU) similarity.

\subsection{In-Model Context Ensembling}
Another interesting feature of our approach that it allows for a
computationally efficient form of ensembling at evaluation time.
First we extract context features $\{f_i\}$ for $k$
large crops in the image.
In our case we used the whole image, $80\%$
size squares in each corner and one same sized square at the center of the
image.
After context features $f_i$ for each of those $k=6$ features
extracted, the final score will be given by $\sum C(f_i, N(p)) / k,$ which
is the average of the combiner classifier $C$ scores evaluated for each pair of
context and object.
This results in a modest ($0.005$-$0.01$ mAP),
but consistent improvement at a relatively small additional cost,
if there are a lot of proposals for each image and the combiner classifier
is much cheaper to evaluate than extracting the features.

\subsubsection{Training Methodology}
All three models: the MultiBox, the context and post-classifier were trained
with the Google DistBelief~\cite{dean2012large} machine learning system using stochastic gradient descent.
The context and post-classifier networks reported in this paper had been
pretrained on the $1.28$ million images of the ILSVRC classification challenge
task. We used only the classification labels during pretraining and ignored any
available location information. The pretraining was done according to the prescriptions
of~\cite{szegedy2015rethinking}. All other models were trained with AdaGrad.
There were two major factors that affected the performance of our models:
\begin{itemize}
\item The ratio of positives versus negatives during the training of the
post-classifier. A ratio of $7:1$ negatives versus positive samples gave good
results.
\item Geometric distortions like random size and aspect ratio distortions
proved to be crucial, especially for the MultiBox model. We have employed
random aspect ratio distortions of up to $1.4{\times}$ in random (either
horizontal or vertical) directions.
\end{itemize}

\section{Results}
\label{sec:results}

  \subsection{Network architecture improvements}
\begin{figure}[h]
\centering
\includegraphics[width=0.49\linewidth]{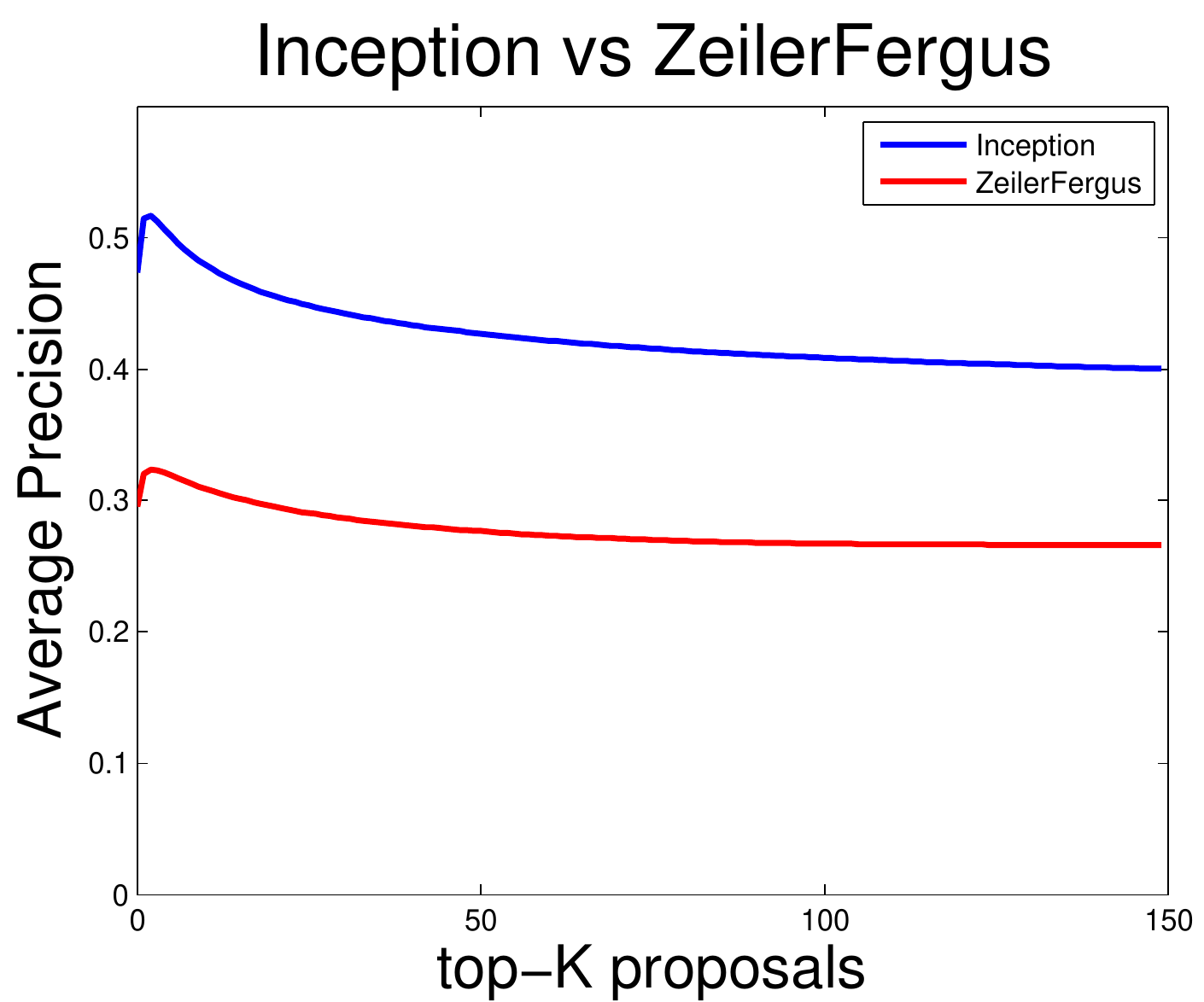}
\includegraphics[width=0.49\linewidth]{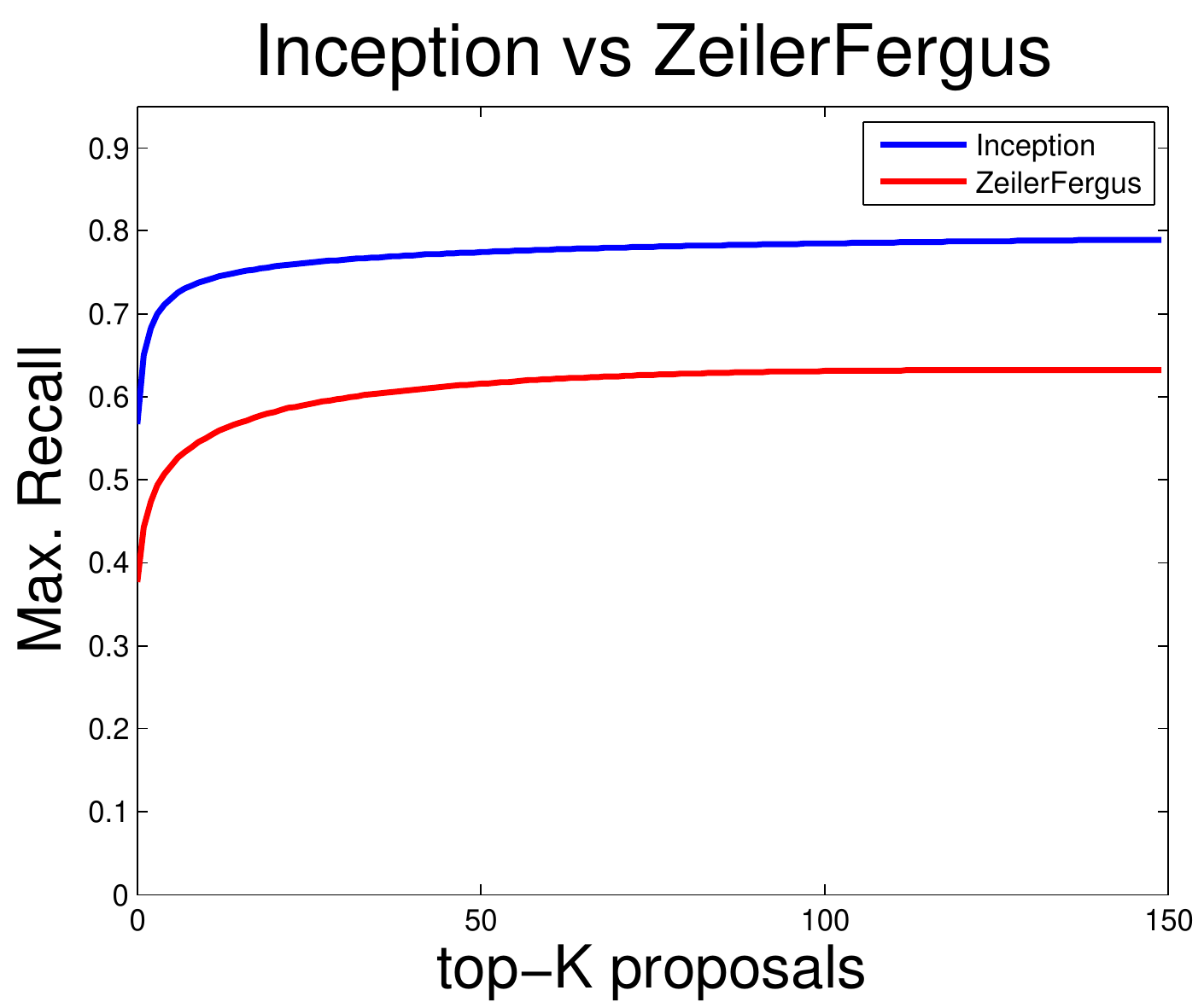}
\caption{The inception architecture is particularly well-suited to localization, drastically improving over the Zeiler-Fergus architecture for MultiBox training.}
\label{fig:inception_vs_zf}
\end{figure}

\begin{figure}[h]
\centering
\includegraphics[width=0.49\linewidth]{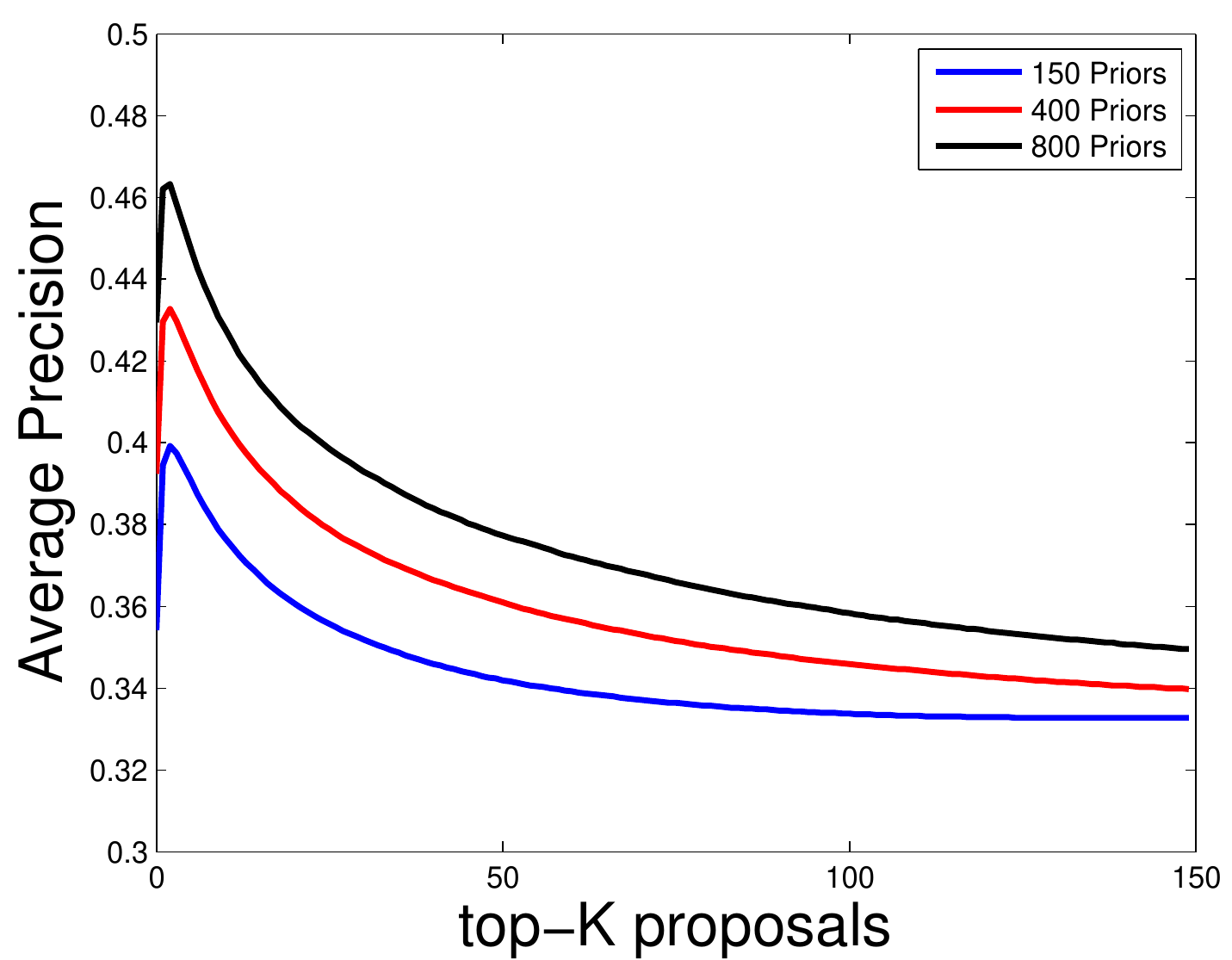}
\includegraphics[width=0.49\linewidth]{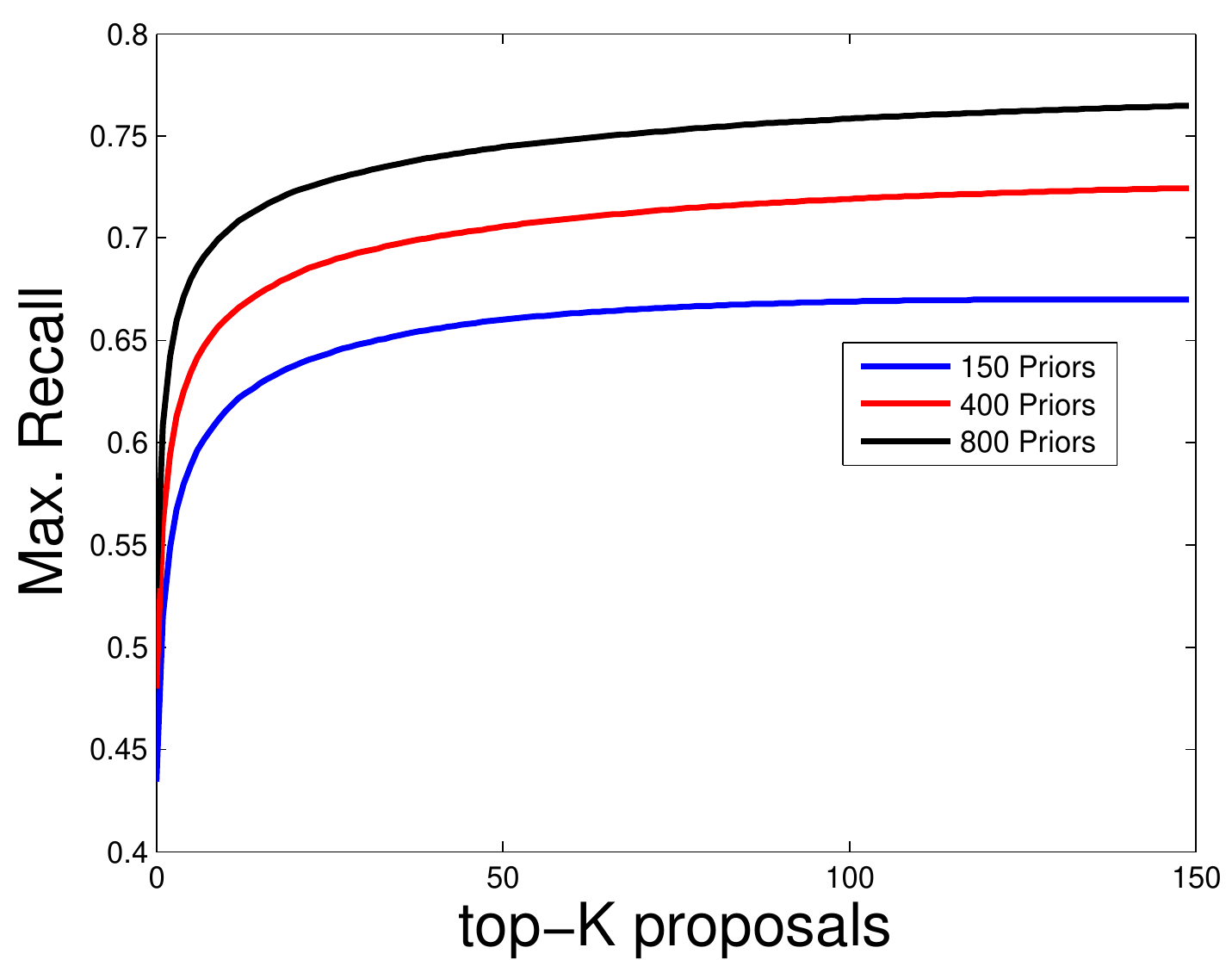}
\caption{Maximum recall and average precision tends to increase as the number of priors is increased.}
\label{fig:numprior_recall_ap}
\end{figure}

In this section we discuss aspects of the underlying convolutional network that benefited the detection performance.
First, we found that switching from a Zeiler-Fergus-style network (detailed in~\cite{zeiler2014visualizing}) to an Inception-style network greatly improved the quality of the MultiBox proposals (see Fig.~\ref{fig:inception_vs_zf}).
A thorough ablative study of the underlying network is not the focus of this paper, but we observed that
for a given budget $K$, both the (class-agnostic) AP and maximum recall increased substantially by the change, as shown in Fig.~\ref{fig:numprior_recall_ap}.
%

Figure \ref{fig:numprior_recall_ap} also shows that with the Inception-style convolutional networks, increasing the number of priors from around 150 (used in the original MultiBox paper~\cite{erhan2014scalable}) to 800 provided a large benefit. Beyond 800, we did not notice a significant improvement.

\subsection{Runtime-quality trade-off}
In this section we present an analysis of the runtime-quality trade-off in our proposed method. The detection runtime is determined mostly by the number of network evaluations, which scales linearly with the number of proposal boxes. Since MultiBox scores the region proposal boxes, we can achieve the maximum quality with the number of network evaluations we can afford by only evaluating the top-$K$ most confident ones.

\begin{figure}
\centering
\includegraphics[width=0.8\linewidth]{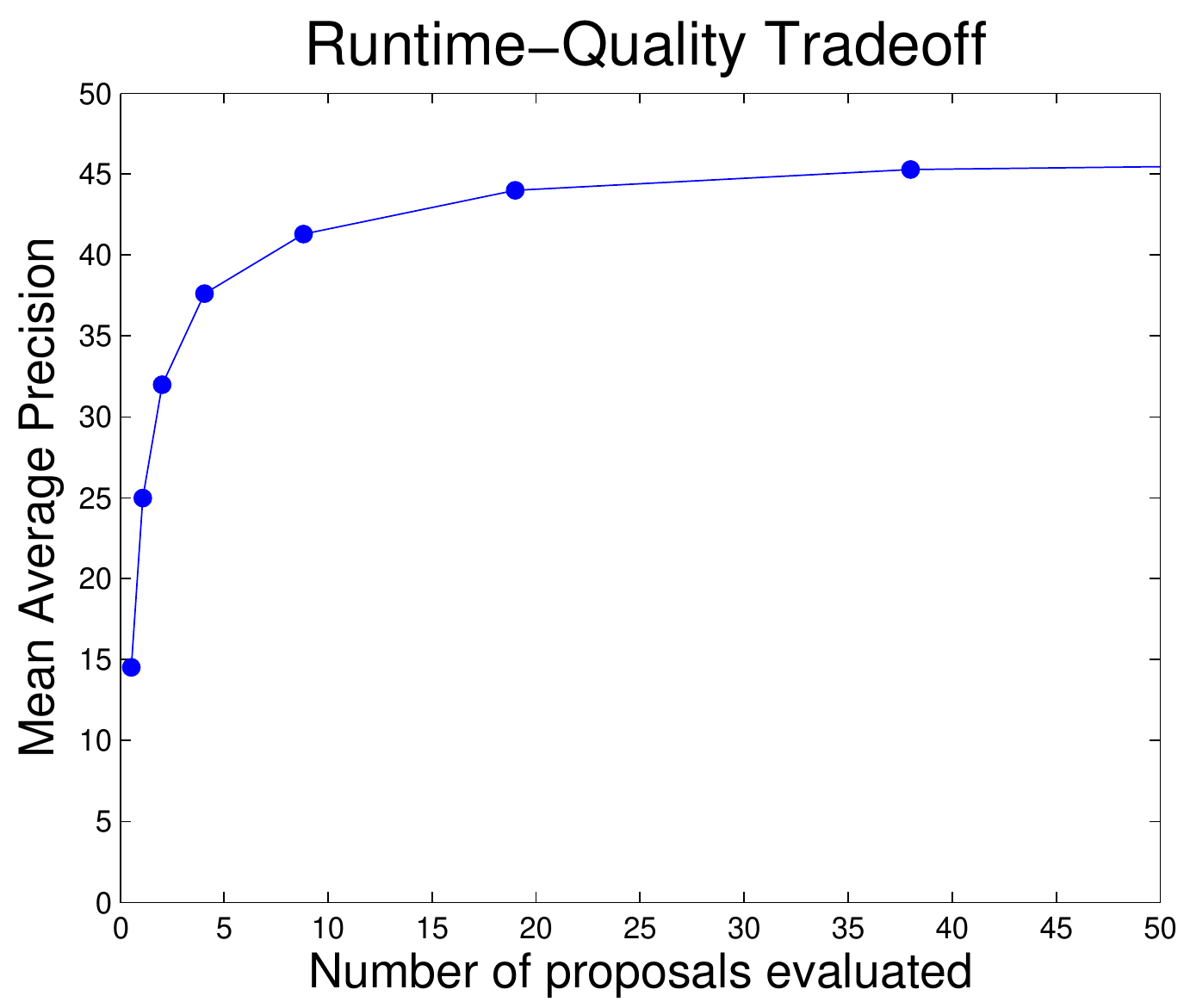}
\caption{A visualization of the trade-off between runtime (scaling with the number of proposal boxes) and quality (mAP) for a single model, single-crop MultiBox model. Note that the network was evaluated once on the image mapped affinely to the network input to generate all proposals using a single network evaluation.}
\label{fig:runtime_quality}
\end{figure}

Figure~\ref{fig:runtime_quality} shows that performance degrades very gracefully with computational budget. Compared to the highest-quality operating point\footnote{The mAP leveled off at around $45.8\%$.}, very competitive performance (e.g. maintaining $> 90\%$ of the mAP) can be achieved with an order of magnitude fewer network evaluations. Also worth noting is that quality does not increase indefinitely with the number of proposals; swamping the post-classifier with low-quality proposals actually reduces the quality.

\subsection{Contextual features}
\label{sec:ilsvrc_results}

We used the same networks to generate both the contextual and non-contextual features,
but the non-contextual network was trained without the extra context features.
Both have a softmax classifier at the top and neither of them used hard negative
mining, they were both pre-trained on the ImageNet classification challenge
and used the same $42$ layers deep Inception variant as the MultiBox
proposal generation model. Table \ref{tab:ilsvrc2014} shows that adding contextual
features greatly improves results.

\begin{table}[h]
\centering
\begin{tabular}{|l|c|}
	\hline
	\textbf{Model} & \textbf{mAP} \\
	\hline
	Non-contextual MSC-MultiBox & $0.473$  \\
        \hline
	Contextual MSC-MultiBox as in fig~\ref{fig:combiner} & $0.5$ \\
	\hline
\end{tabular}
\caption{Control experiments for post-classification using contextual versus non-contextual models. Both models were trained as on multi-scale convolutional MultiBox-based proposals and with the multi-crop methodology described below.}
\label{tab:ilsvrc2014}
\end{table}

\subsection{Multibox on many image crops}

The most efficient MultiBox solution generates proposals from a single network evaluation on a single image crop.
We can increase the quality at the cost of a few more network evaluations by
taking multiple crops of the image at multiple scales and locations, and
combining all of the generated proposals and applying non-maximal suppression.

In the MultiBox case, one needs to be cautious: if the
proposals are kept indiscriminately, then the system will produce high
confidence boxes from partial objects that overlap the crop. This naive
implementation ends up with a loss of quality.
Our solution was to drop all the proposals that are not completely
contained in the $(0.1, 0.1) - (0.9, 0.9)$ sub-window of the crop.
However this implies that MultiBox should be applied on highly overlapping windows.
We have run two experiments in which a $299\times 299$ crop
was slid over the image
such that each window overlaps at least $50\%$ (or $62.5\%$) each of its
neighboring window in the dimension they are adjacent, respectively.
This allows enough room for small object to be picked up by at least one of
the crops evaluated with MultiBox.

Table~\ref{tab:multi_crop} demonstrates that we can get almost $5\%$ mAP
improvement by taking multiple image crops in the proposal generating step.
The resulting number of proposals increases from $13$ per image to $51$ per image
on average, but is still significantly lower than that used by Selective Search.

\subsection{ILSVRC2014 detection challenge}

In this section we combine multi-scale convolutional MultiBox
proposals with context features and a post-classifier
network on the full 200-category ILSVRC2014 detection challenge data set.
\begin{table}
\centering
\begin{tabular}{|l|c|c|}
	\hline
	\textbf{Model} & \textbf{mAP} & \textbf{boxes} \\
	\hline
	MSC-MultiBox single-crop & $0.45$ & $13$ \\
	\hline
	MSC-MultiBox multi-crop (0.625) & $0.5$ & $51$ \\
	\hline
\end{tabular}
\caption{Control experiments using our models on the ILSVRC 2014 detection challenge validation set with a single model.}
\label{tab:multi_crop}
\end{table}

Table~\ref{tab:ilsvrc2014} shows several rows, each of which lies on a
different point along the runtime-quality trade-off.
Note that our improved MultiBox pipeline with a \emph{single} crop
yields $0.45$ mAP, which exceeds last year's GoogLeNet ensemble
validation performance in the ILSVRC2014 competition,
and is even higher than the latest and best known result published with
Deep-ID-Net~\cite{ouyang2014deepid}.
In addition, we attain superior performance at the high-precision operating point.
Given a single Multibox region proposal network and a single post-classifier model,
we obtain $0.499$ mAP.

We obtain even better results by using an ensemble of models. Naive ensembling,
such as the one done by the GoogLeNet team on the ILSVRC 2014 detection
challenge~\cite{szegedy2014going}, uses a single
Multibox network to propose boxes and then averages the result of several
post-classifier models on the boxes. When we tried this with 3 post-classifier
models, we got a mAP of $0.506$ -- a slight improvement.

We wanted to leverage the results of several different Multibox models, as well.
Intuition suggests that box proposals that are consistent across several different
Multibox models are more likely to be high-quality proposals. To capture this,
we designed the following ensembling approach for $N$ Multibox models.
For the boxes of each Multibox model $j \in [1, N]$, we can use either a single
post-classifier model, or average the scores of several post-classifier models,
obtaining a set of bounding boxes $l_i^j$ and class scores $c_{i,k}^j$,
for each class $k$, and post-classifier model $i$. For each class score
$c_{i,k}^j$, we aggregate scores from the other Multibox models as follows:
\begin{align}
  s_{i,k}^j = \frac{1}{N} \cdot (c_{i,k}^j + \sum_{n \neq j} {\max_m (J(l_i^j, l_m^n) \cdot c_{m,k}^n)},
\end{align}
where $J(\cdot)$ is the Jaccard overlap between the bounding boxes. Put in words,
the objective above reinforces detections that have consistent matches in the
other Multibox results both in terms of location (high Jaccard overlap) and high
score. After computing these scores for all detections and scores of all Multibox
models, we apply non-max suppression to keep only the best ones. This ensembling approach
yielded $0.52$ mAP with two Multibox models, a substantial improvement over the naive version.
\begin{table}[h]
\centering
\begin{tabular}{|l|c|}
	\hline
	\textbf{Model} & \textbf{mAP (\%)}\\
	\hline
	Deep Insight ensemble & $0.41$ \\
	\hline
	GoogLeNet ensemble & $0.44$  \\
	\hline
	DeepID-Net ensemble & $0.44$ \\
	\hline
	MSC-MultiBox single-crop & $0.45$ \\
	\hline
	MSC-MultiBox multi-crop, one model & $0.5$ \\
	\hline
	Ensemble of two models of MSC-MultiBox & $0.52$  \\
	\hline
\end{tabular}
\caption{Comparison to the existing state-of-the-art results~\cite{russakovsky2014imagenet}.}
\label{tab:compare}
\end{table}
Table~\ref{tab:compare} demonstrates that multi-scale convolutional MultiBox
establishes a new state-of-the-art by a healthy margin.
\begin{table}[h]
\centering
\begin{tabular}{|c|c|c|}
	\hline
        {\bf category} & AP & Recall at 60\% precision \\
        \hline
	person & $0.6$ & $63.1\%$ \\
	\hline
	bird & $0.91$ & $93.1\%$ \\
        \hline
        dog & $0.94$ & $95.7\%$ \\
        \hline
        can opener & $0.55$ & $56.3\%$ \\
	\hline
        table & $0.36$ & $33\%$ \\
	\hline
        horizontal bar & $0.097$ & $0\%$ \\
	\hline
\end{tabular}
\caption{Performance of the two-model ensemble on a few selected classes of ILSVRC-2015.}
\label{tab:selected}
\end{table}
\subsection{Quality of MSC-MultiBox Proposals}
\label{sec:proposalquality}

\begin{table}[h]
\centering
\begin{tabular}{|l|c|c|c|c|}
  \hline
  \multirow{2}{*}
  {\bf proposals} &
  \multicolumn{4}{|c|}{\textbf{Recall at Jaccard overlap}} \\
  & $0.5$ & $0.6$ & $0.7$ & $0.8$ \\ \hline\hline
  2.7 & { 0.2}  & { 0.18} & { 0.14} & { 0.09} \\ \hline
  8.3 & { 0.36} & { 0.31} & { 0.24} & { 0.15} \\ \hline
  22  & { 0.52} & { 0.44} & { 0.34} & { 0.2}  \\ \hline
  55  & { 0.64} & { 0.55} & { 0.42} & { 0.25} \\ \hline
  228 & { 0.77} & { 0.68} & { 0.53} & { 0.31} \\ \hline
  616 & { 0.84} & { 0.75} & { 0.6}  & { 0.35} \\ \hline
  947 & { 0.86} & { 0.78} & { 0.64} & { 0.37} \\ \hline
  2056 & {0.9} & { 0.83} & { 0.68} & { 0.4} \\ \hline
  4168 & { 0.92} & { 0.86} & { 0.72} & { 0.42} \\ \hline
  8409 & { 0.93} & { 0.88} & { 0.75} & { 0.43} \\
  \hline
\end{tabular}
\caption{Multi-scale convolutional MultiBox Per-class average recall
  at various Jaccard thresholds on the Microsoft-COCO~\cite{lin2014microsoft}
  validation set. Please refer to the corresponding
  Figure~\ref{fig:agnosticrecall} which shows that MultiBox outperforms MCG at
  overlap thresholds up to $0.75$. It still surpasses the recall of MCG for $0.8$
  when the budget is below $200$ proposals per image.
}
\label{tab:agnosticrecall}
\end{table}

\begin{figure}
\centering
\includegraphics[width=\linewidth]{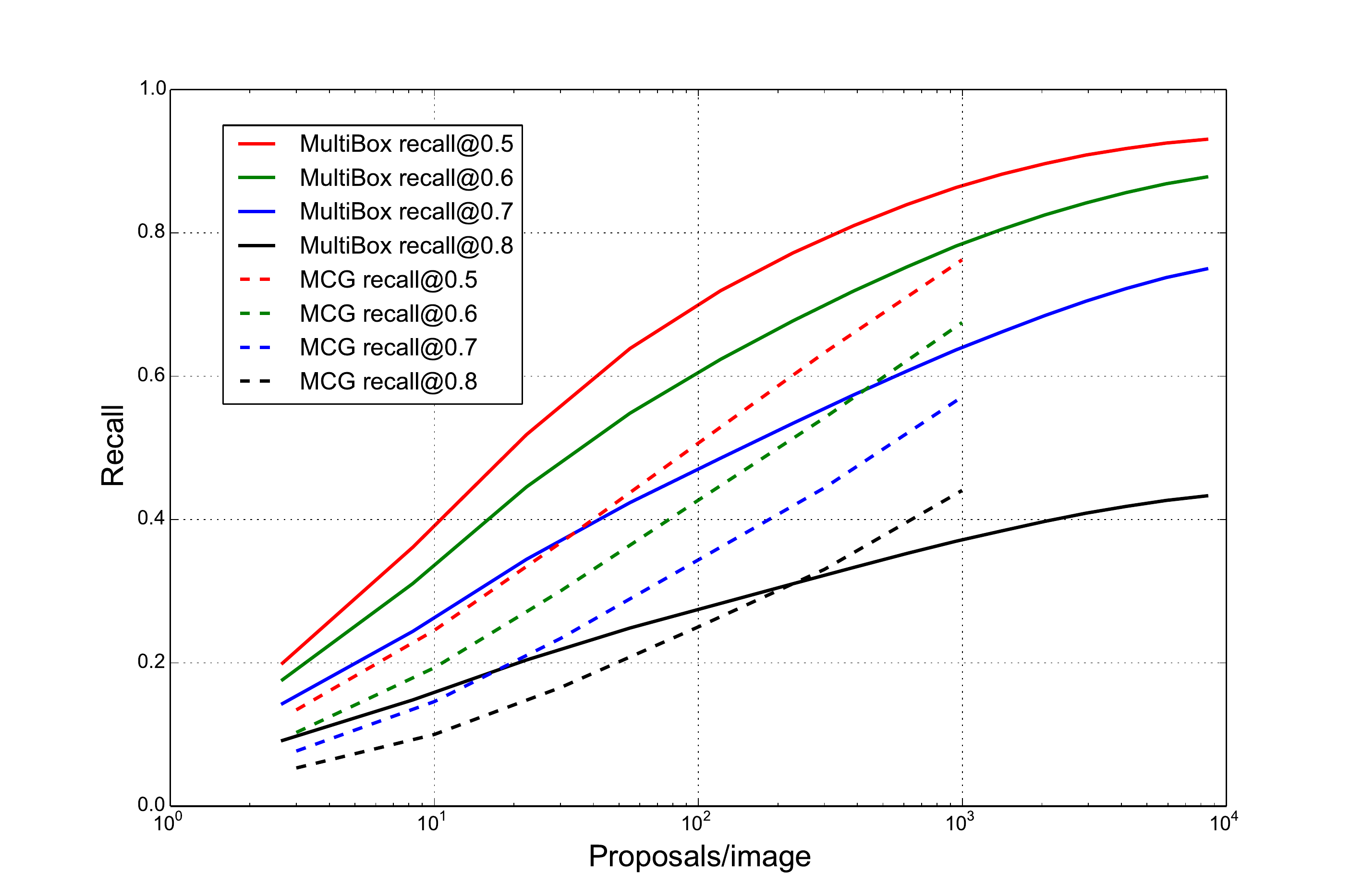}
\caption{Per-class Average Recall of Multi-scale convolutional MultiBox
  at Jaccard ranging between $0.5$ and $0.8$ on the
  Microsoft-COCO~\cite{lin2014microsoft} data set
  compared with proposals generated with MCG\cite{PABMM2015}
  (Multiscale Combinatorial Grouping). The corresponding MultiBox
  recall numbers are reported in Table~\ref{tab:agnosticrecall}.}
\label{fig:agnosticrecall}
\end{figure}

In this section, we are comparing the coverage of our class-agnostic proposal
generation method with the state-of-the-art
Multiscale Combinatorial Grouping~\cite{PABMM2015} approach on the
Microsoft-COCO~\cite{lin2014microsoft} validation set.

For this purpose, we have trained a class-agnostic MultiBox model on top of
the Inception-v3 network~\cite{szegedy2015rethinking} using the
TensorFlow~\cite{tensorflow2015-whitepaper} large scale distributed system
with asynchronous gradient descent with 30 model replicas for
2 million batches, each of size $32$.

For MultiBox, we have evaluated the crops from each image at three scales:
\begin{itemize}
  \item The whole image was warped to the $299\times299$ receptive field of
    the network.
  \item A $299\times 299$ square crop was slid on the image such that the
    minimum overlap between adjacent crops is at least $0.5$. Only those
    proposals are kept that are completely contained in the center
    square covering $0.8\times 0.8$ of the crop.
  \item A $185\times 185$ square crop was slid on the image such that the
    adjacent crops have at least $0.5$ overlap. This crop is scaled up to
    the $299\times 299$ receptive field. Again all predicted proposals
    not fully contained in the the center $0.8\times 0.8$ square are ignored.
\end{itemize}

Finally, for each image, we took the union of all proposals from each crop
and ran non-maximum-suppression with Jaccard threshold $0.85$.

To compute recall, the proposals are ranked by their confidence scores. We
have took 15 different pre-sigmoid score thresholds ranging from $2$ to $-12$.
which gave rise to various average numbers of proposals per image. The results
are reported in Table~\ref{tab:agnosticrecall} and the corresponding
Figure~\ref{fig:agnosticrecall}.

As one can see, MultiBox significantly outperforms MCG below 2000 proposals,
especially for lower overlap threshold. MCG only outperforms MultiBox at $0.8$
or higher thresholds with over $300$ proposals. However, we expect that
MultiBox might do better if pre-processed with less aggressive
Non-Maximum-Suppression threshold (exceeding the currently used $0.85$
threshold) when optimizing for recall at tight thresholds (above $0.75$).

\section{Conclusions}
In this work we demonstrated a method for high-quality object detection that is simple, efficient and practical to use at scale.

The proposed framework flexibly allows the choice of operating point along the runtime-quality trade-off curve.
Even using single-crop multi-scale convolutional MultiBox with only several dozen proposals per image on average, we exceed the previously-reported state-of-the-art ILSVRC2014 detection performance,
outperforming even highly-tuned ensembles using costly Selective Search proposal generation.
At the high-quality end of the curve, we outperform the nearest reported mAP by over $10\%$ relative.

We conclude that learning-based proposal generation has closed
the performance gap with state-of-the-art engineered proposal
generation methods, MCG~\cite{PABMM2015} in our study,
while reducing the computational cost of detection.
This is mostly the result of improved underlying network
architecture especially the use of multi-scale convolutional proposal
generation.
Improvements in training methodology, context modeling and
inference-time tricks like multi-crop evaluation and
in-model ensembling resulted in modest, but significant cumulative
gains on ILSVRC Detection 2014. Multi-scale convolutional MultiBox
is not just a computationally more efficient replacement for static
proposal generating algorithms; by providing a smaller number of
higher-quality proposals, multi-scale convolutional MultiBox improves
the overall object detection performance.

{\small
\bibliographystyle{ieee}
\bibliography{references}
}

\end{document}